# Refinment and Coarsening of Bayesian Networks

**Kuo-Chu Chang and Robert Fung**
Advanced Decision Systems
1500 Plymouth Street
Mountain View, California    94043-1230

## Abstract

In almost all situation assessment problems, it is useful to dynamically contract and expand the states under consideration as assessment proceeds. Contraction is most often used to combine similar events or low probability events together in order to reduce computation. Expansion is most often used to make distinctions which have significant probability in order to improve the quality of the assessment. Although other uncertainty calculi, notably Dempster-Shafer [4], have addressed these operations, there has not yet been any approach of refining and coarsening state spaces for the Bayesian Network technology.

This paper presents two operations for refining and coarsening the state space in Bayesian Networks. We also discuss their practical implications for knowledge acquisition.

## 1 Introduction

Bayesian Networks [1], [2] is a technology for reasoning under uncertainty and has been used primarily to address situation assessment problems (e.g., medical diagnosis, battlefield assessment). In situation assessment, the problem is to infer the strength-of-belief (i.e., probability) in certain propositions given a set of internal beliefs (e.g., rules) and external evidence. In general, the evidence about a situation does not come in all at once, instead it is received over a period of time. As evidence is received and beliefs are updated, some distinctions which were previously irrelevant become relevant and some distinctions which were previously relevant become irrelevant. In general, it will be infeasible to consider all possible relevant distinctions throughout the assessment process due to resource limitations. Therefore an opportunistic approach is needed in which new states can be added and existing states which are similar or insignificant can be combined or removed dynamically as the assessment proceeds. Other uncertainty calculi have also recognized the importance of this problem, notably Dempster-Shafer [4].

The introduction of new distinctions to a state space *refines* the state space and the removal of distinctions *coarsens* the state space. These operations must fulfill the intuitive constraint that their use must not affect beliefs which do not directly involve the refined or coarsened state space. This paper presents operations for refining and coarsening the state spaces of Bayesian Networks. The inputs to the operations are a target node and the desired refinement or coarsening. The outputs of the operation are revised conditional probability distributions for the target node and for the target node's successors which correspond to the modified state space of the node.

There are three important observations about these operations. First, to satisfy the constraint that refinement and coarsening operations do not affect the probability of states not involved in the operation, it is sufficient that the operations do not affect the probability of states in the "neighborhood" of the node under consideration. It can be easily shown that this "neighborhood" of a node is the Markov boundary of the node, namely, the node's predecessors, successors, and successors' predecessors. In other words, if the joint probability distribution of the blanket (other than node itself) is not changed by the operation, then the joint probability distribution of the entire network (other than the node itself) will also be unchanged by the operation.

Second, since refinement operations "introduce" information to the network some judgements need to be made about the relative weights of the new distinctions. This can be done by modifying the relationships (*i.e.,* the conditional probabilities) between the refined variable, its predecessors, and its successors. In order to satisfy the Markov boundary condition described above, certain constraints need to be met in modifying these relationships. The degree of freedom in assigning the new probabilities is limited.

Third, while coarsening operations can always be exact (i.e., satisfy the Markov boundary condition), the associated costs are high enough that it may be desirable to make the operation approximate. In such circumstances, information may be lost due to the approximation. By the loss of information, we mean that the resulting network will have a different probability distribution than the original one. However, if the states to be coarsened are "similar" enough, the resulting impact will be small.

This paper is organized as follows. Section 2 describes the refinement and coarsening operations. Section 3 presents some detailed examples. Some discussions and

476

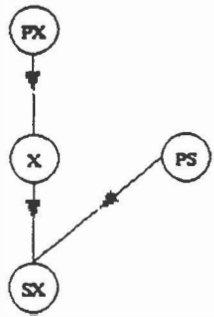

Figure 1: A Simplified Network

concluding remarks are given in Section 4.

## 2 The Refinment and Coarsening Operations

In this section, two sets of related operations, one called "external" and one called "internal", for refining and coarsening a node's state space are presented. The node to be refined or coarsened will be referred to as the "target" node. First set of operations is called "external" since a new node is added "externally" to the target node which is a successor to the target node and whose state space is the desired modified state space. In the "internal" operation, however, the operation works "internally" on the node without changing the topology of the original network.

The external operation is straight-forward. An external node is added to the diagram which has no successors and has the target node as its only predecessor. The state space of the external node is the desired refinement or coarsened state space of the target node. The arc (conditional probabilities) between these two nodes describe the mapping, either refinement or coarsening, between their state spaces. The target node is then removed from the graph probabilistically based on the arc reversal and node removal operations [3]. This leaves the external node in place of the target node in the new graph. By doing so, an extra arc is introduced between the predecessors and the successors of the target node. The advantage of this approach is its simplicity. The disadvantage of the approach is the change in the network topology. In a dense network, this operation may introduce many extra arcs.

An example of the operation is shown in Figures 1 and 2. Figure 1 shows the original network. Suppose that $x$ is the target node. We first add an external node $x1$ as the original node's successor with the desired new state space. Figure 2 shows the resulting network after the removal of the target node $x$. As can be seen, an extra arc has been introduced between the predecessor and the successor of $x$.

The internal operation refines and coarsens the state space of a node without changing the topology of the

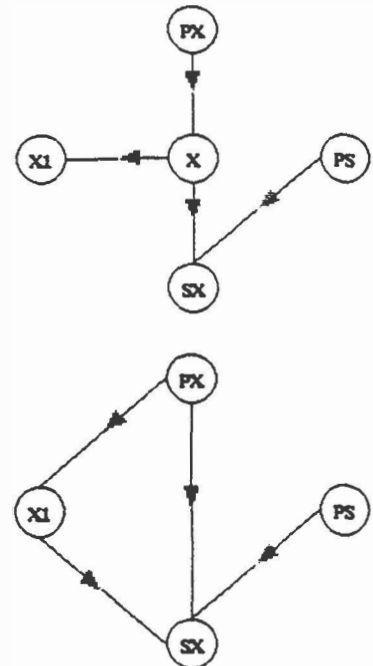

Figure 2: Modifying Network with External Operation

networks. This operation takes four inputs:

- a state node ($x$) whose state space ($\Omega_x$) is to be refined or coarsened,

- a new state space ($\Omega'_x$),

- a relationship between ($\Omega_x$) and ($\Omega'_x$) which specifies which values $\omega_x$ in $\Omega_x$ are refined or coarsened into which values $\omega'_x$ in $\Omega'_x$.

- as auxiliary information, the Markov blanket of the node. The Markov blanket of $x$ requires that the state spaces of it's predecessors $P_x$ and successor's predecessors $P_{s_x}$ be specified as well as the probability distributions of it's successors $S_x$ (see Figure 1).

The output of the operation are two sets of probability distributions:

- the new conditional distribution for $x$, $p'(x|P_x)$ and

- modified distributions for the successors of $x$, $p'(s_x|x, P_{s_x})$

We specify the relationship between $\Omega_x$ and $\Omega'_x$ with two mappings. The refinement mapping $Refine$ maps a single value in $\Omega_x$ into multiple values in $\Omega'_x$ and the coarsening mapping $Coarsen$ maps a single value in $\Omega'_x$ into multiple values in $\Omega_x$.

In the refinement operation, for those values $\omega_x$ in $\Omega_x$ which are refined into $\omega'_x \in Refine(\omega_x)$ in $\Omega'_x$, an obvious constraint of the new probability distribution is

$$p(\omega_x|P_x) = \sum_{\omega'_x \in R(\omega_x)} p(\omega'_x|P_x) \quad (1)$$

Since $\omega'_x$ does not have to be equally weighted, one needs to make the judgements about the relative weights of the new distribution.

The Markov boundary of the state node $x$ includes $P_x$, $S_x$, and $P_{S_x}$. They "shield" the node $x$ from the rest of the network. It can be easily shown that if the joint probability distribution of the Markov boundary is not affected by the refinement operation, then the rest of the network will not be affected. To keep the joint probability distribution of the Markov boundary the same before and after the refinement operation, the constraint to be satisfied is chosen as $p(S_x|P_x, P_{S_x})$, namely,

$$
\begin{aligned}
p(S_x|P_x, P_{S_x}) &= \sum_x p(s_x|x, P_{s_x})p(x|P_x) \\
&= \sum_{x'} p(s_x|x', P_{s_x})p(x'|P_{x'})
\end{aligned}
\tag{2}
$$

In other words, for the value $\omega_x$ to be refined,

$$
p(s_x|\omega_x, P_{s_x})p(\omega_x|P_x) = \sum_{\omega'_x \in R(\omega_x)} p(s_x|\omega'_x, P_{s_x})p(\omega'_x|P_x)
\tag{3}
$$

need to be satisfied for all values of $P_x$. An obvious solution satisfies the above constraints regardless of $p(\omega'_x|P_x)$ is,

$$
p(s_x|\omega'_x, P_{s_x}) = p(s_x|\omega_x, P_{s_x})
\tag{4}
$$

This solution states that, regardless of how the conditional probabilities $p(\omega'_x|P_x)$ being assigned, as long as they satisfy eqn. (1), then if the conditional probabilities of the successors $S_x$ given the refined values are set to be the same as that of the original value, then the constraint (3) is satisfied. This solution allows us to assign arbitrary proportions in the *upper* arc $p(\omega'_x|P_x)$ but leaves no freedom in determining the *lower* arc $p(s_x|\omega'_x, P_{s_x})$.

In general, the above solution may be too restrictive. In fact, if the proportions $p(\omega'_x|P_x)$ assigned in eqn. (1) are the same for all the predecessor values, namely, if

$$
\frac{p(\omega'_x|P_x)}{\sum_{\omega'_x \in R(\omega_x)} p(\omega'_x|P_x)} = K(\omega'_x)
\tag{5}
$$

where $K(\omega'_x)$ is a function depending only on $\omega'_x$, then eqn.(3) can be reduced to a single constraint, i.e.,

$$
p(s_x|\omega_x, P_{s_x}) = \sum_{\omega'_x \in R(\omega_x)} p(s_x|\omega'_x, P_{s_x})K(\omega'_x).
\tag{6}
$$

In this case, $p(s_x|\omega'_x, P_{s_x})$ do not need to be the same as $p(s_x|\omega_x, P_{s_x})$ and as long as they satisfy the constraint (6), we have freedom to assign their numbers. In other words, by imposing one more restriction (5) in obtaining the upper arc, more freedom is allowed in choosing the lower arc.

In the internal coarsening operation, for those values $\omega_x$ in $\Omega_x$ which are coarsened, two constraints similar to (1) and (3) need to be satisfied,

$$
p(\omega'_x|P_x) = \sum_{\omega_x \in C(\omega'_x)} p(\omega_x|P_x)
\tag{7}
$$

and

$$
p(s_x|\omega'_x, P_{s_x}) = \frac{1}{p(\omega'_x|P_x)} \sum_{\omega_x \in C(\omega'_x)} p(s_x|\omega_x, P_{s_x})p(\omega_x|P_x)
\tag{8}
$$

If both of these constraints are satisfied, then the coarsening procedure is exact and the rest of the network will not be affected. However, if no single value of $p(s_x|\omega'_x, P_{s_x})$ can be found to satisfy (8) for all values of $P_x$, then that means those $\omega_x$ in $C(\omega'_x)$ cannot be coarsened without changing the joint probability of the network. In other words, some information may be lost when "aggregating" those state values together and the new network will be "inconsistent" with the old one. If that is desirable, one can either use the external operation described earlier or use the internal operation with some approximation. If the values to be coarsened are "similar", namely, the values of $p(s_x|\omega'_x, P_{s_x})$ calculated based on the right hand side of (8) with different values of $P_x$ are close, then the approximation will have small impact on the rest of the graph. A reasonable approximation under such situation will be to calculate $p(s_x|\omega'_x, P_{s_x})$ as the average of the values obtained from (8).

## 3 Illustration of the Operations

We illustrate the refining and coarsening operations for both the external and internal approaches with the following examples. First consider the graph given in Figure 3. In this example, the root node $M$ has two values, Military Unit Type A and Type B. The second node $V$ has two values representing whether a vehicle exists in a particular place and time. The terrain condition node $T$ has two values, good and bad. The feature node $F$ has two predecessors, vehicle $V$ and terrain condition $T$, and has three values feature A, feature B, and feature of Others. The original probability distribution of the graph as well as the computed posterior probabilities of each node given the evidence are also given in Figure 3.

In this example, suppose we are only interested in distinguishing whether there is a vehicle or not, which can help us identifying the type of military unit. When the posterior probability of the presence of vehicle becomes very high as supported by evidence, we may become interested in more details about the vehicle. Suppose, we are interested in what type of vehicle it is, first we refine the state value $Y$ of node $V$ into two values, Tank $A$ and Truck $U$. With the external operation, we can add an artificial node $V1$, in which tank and truck are split, say in a one to four ratio (see Figure 4). After removing the original node $V$, the resulting graph, the corresponding conditional probabilities and the posterior probabilities of each node are also shown in Figure 4.

With the internal approach, first we assign probabilities for the upper arc. As in the external approach, we split vehicle into tank and truck with one to four ratio and we assume it is independent of military unit type. The new conditional probabilities of refined node $V$ given $M$ is shown in Figure 5. In this case, the condition given in eqn.(5) is satisfied, we therefore have freedom in assigning the conditional probabilities of the lower arc as long as they satisfy the constraint given in eqn. (6),

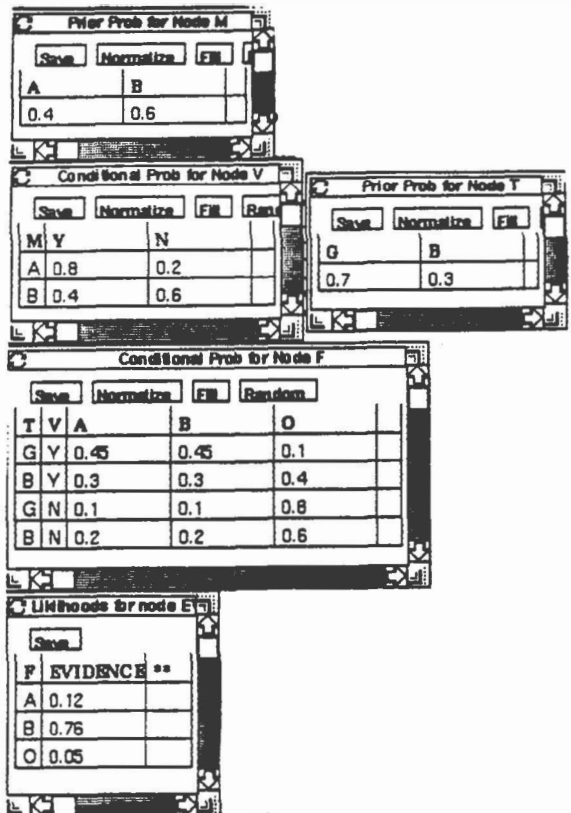

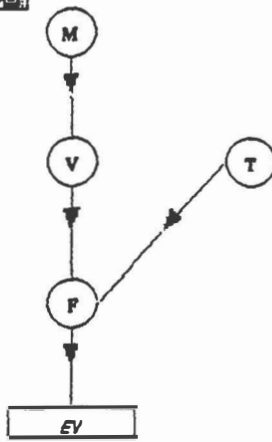

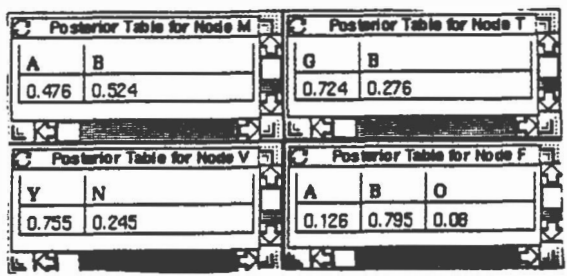

Figure 3: An Example Network

namely,

$$p(F_i|V_y, T_j) = \frac{[p(F_i|V_a, T_j)p(V_a|M_k) + p(F_i|V_u, T_j)p(V_u|M_k)]}{p(V_y|M_k)} \quad (9)$$

where $F_i$ is the $i-th$ value of node $F$. It can be seen that this is a line equation in a two-dimensional space. For instance, for $F_i = a$, $T_j = g$ we have,

$$p(F_a|V_y, T_g) = 0.45 = 0.2p(F_a|V_a, T_g) + 0.8p(F_a|V_u, T_g) \quad (10)$$

Therefore any pair of values $p(F_a|V_a, T_g)$ and $p(F_a|V_u, T_g)$ between 0.0 and 0.9 (because $p(F_a|V_i, T_g) = 0.1$) and satisfy eqn. (10) are feasible. Based on the constraints, we choose the feasible conditional probabilities of node $F$ as given in Figure 5. The idea of choosing those numbers is that given vehicle is a Tank, the probability of detecting feature $A$ is much higher than detecting feature $B$. On the other hand, there is a slightly higher probability to detect feature $B$ than feature $A$ from Truck. With these new arcs, the posterior probabilities of each node given the evidence are shown in Figure 5. As can be seen, other than node $V$, the probabilities are the same as the one in Figure 4. Apparently, because of the new arcs and because the evidence favor feature $B$, the new posterior probability of Tank is smaller.

We may also choose the upper arc in such a way that the split of vehicle between tank and truck depends on the military unit. For example, as given in Figure 6, the percentage of tank in type $A$ military unit is assumed to be much more than that in type $B$ military unit. In this case, the condition given in eqn. (5) is not satisfied, the only solution that can satisfy eqn. (6) is the obvious solution given in eqn. (4), namely, the conditional probabilities of node $F$ given the refined values $V_a$ and $V_u$ must be the same as that of the original value $V_y$ as shown in Figure 6. The resulting posterior probabilities also given in Figure 6 show visible changes in node $V$. Note that, while it is possible in refinement to have different ratio of splitting in the upper arc with the internal approach, it can not be done using the external operation. As shown in Figure 4, the external operation always produces the same ratio of splitting in the upper arc which may not be desirable in certain cases.

With the refined network given in Figure 5 or Figure 6, if we coarsen the state values $V_a$ and $V_u$ back into $V_y$ using the internal operation, obviously, the results will be the same as the one in Figure 3. However, in many cases, no coarsening can be done without changing the joint probability distribution of the network. For instance, if the conditional probabilities of the same network is set to be the one given in Figure 7, then no single value of $p(F_i|V_j, T_k)$ can be found to satisfy (8) for all values of $M_i$. That means we either have to approximate the coarsening or we can use the external operation. The approximated values we use for $p(F_i|V_j, T_k)$ is to take average of the values obtained from (8) as described in the previous section. The resulting conditional probabilities between $F$ and $V$ and the computed posterior probabilities are also given in Figure 7. The results of external operation which combine $V_a$ and $V_u$ into $V_y$ are shown in

**Conditional Prob for Node V1**

Save  Normalize  Fill  Random

| V | A | U | N |
|---|---|---|---|
| Y | 0.2 | 0.8 | 0.0 |
| N | 0.0 | 0.0 | 1.0 |

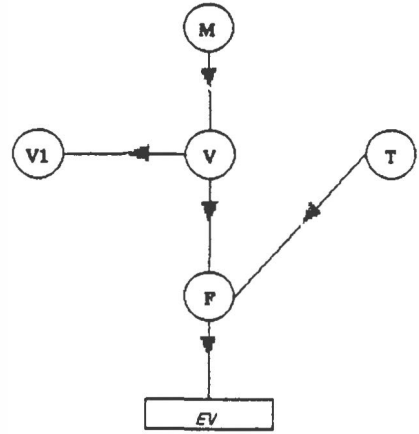

**Conditional Prob for Node F**

Save  Normalize  Fill  Random

| M | V1 | T | A | B | O |
|---|----|---|-----|-----|-----|
| A | A | G | 0.45 | 0.45 | 0.1 |
| B | A | G | 0.45 | 0.45 | 0.1 |
| A | U | G | 0.45 | 0.45 | 0.1 |
| B | U | G | 0.45 | 0.45 | 0.1 |
| A | N | G | 0.1 | 0.1 | 0.8 |
| B | N | G | 0.1 | 0.1 | 0.8 |
| A | A | B | 0.3 | 0.3 | 0.4 |
| B | A | B | 0.3 | 0.3 | 0.4 |
| A | U | B | 0.3 | 0.3 | 0.4 |
| B | U | B | 0.3 | 0.3 | 0.4 |
| A | N | B | 0.2 | 0.2 | 0.6 |
| B | N | B | 0.2 | 0.2 | 0.6 |

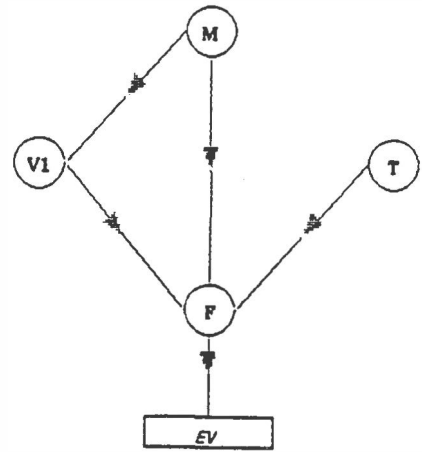

**Conditional Prob for Node V1**

Save  Normalize  Fill  Random

| M | A | U | N |
|---|------|------|-----|
| A | 0.16 | 0.64 | 0.2 |
| B | 0.08 | 0.32 | 0.6 |

**Likelihoods for node F**

Save

| F | EVIDENCE |
|---|----------|
| A | 0.12 |
| B | 0.76 |
| O | 0.05 |

**Posterior Table for Node M**

| A | B |
|-------|-------|
| 0.476 | 0.324 |

**Posterior Table for Node**

| G | B |
|-------|-------|
| 0.724 | 0.276 |

**Posterior Table for Node V1**

| A | U | N |
|-------|-------|-------|
| 0.151 | 0.604 | 0.245 |

**Posterior Table for Node F**

| A | B | O |
|-------|-------|------|
| 0.126 | 0.795 | 0.08 |

Figure 4: Refined Network with External Operation

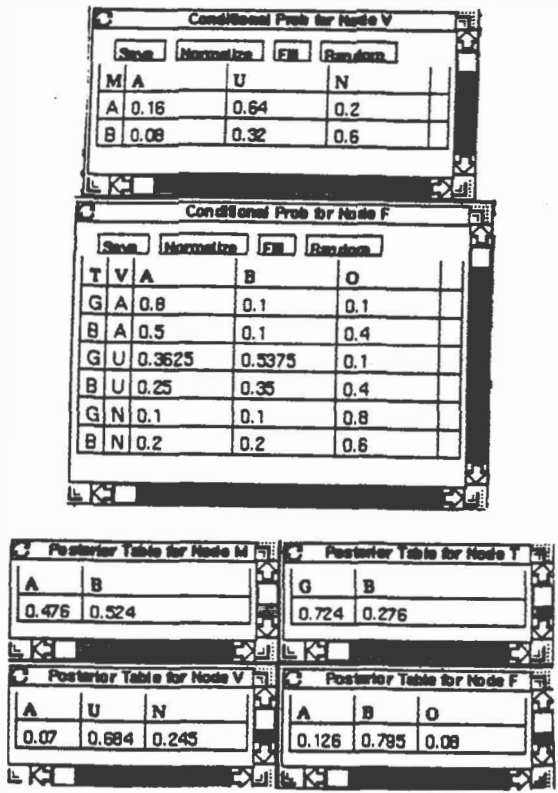

Figure 5: Refined Network with Internal Operation I

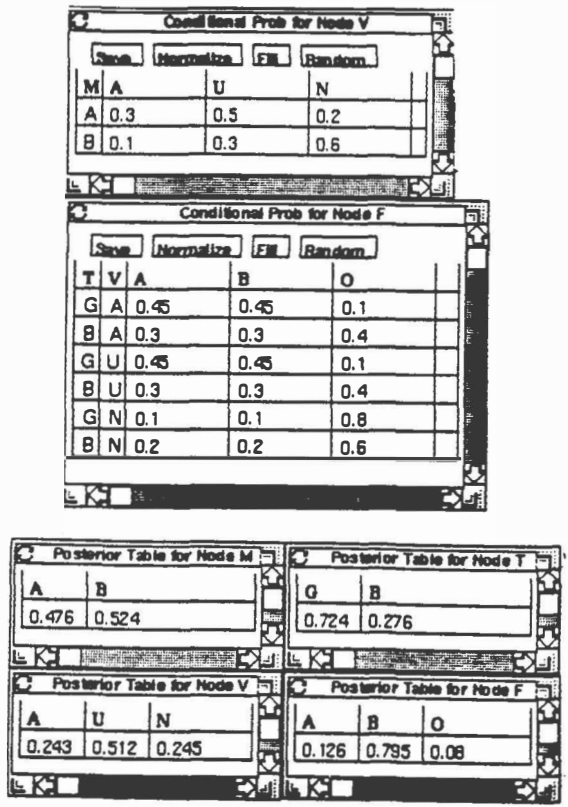

Figure 6: Refined Network with Internal Operation II

Figure 8. As can be seen, the results in Figures 7 and 8 are very similar, except in the case of Figure 8, we have introduced an extra arc between node $F$ and $M$.

## 4 Discussion and Conclusion

This paper has presented two operations, the external operation and the internal operation for refining and coarsening the state spaces of nodes in Bayesian Networks. The operations satisfy the constraint of leaving unrelated probabilities in the network unchanged. Through the refining or coarsening operations, one can "emphasize" the important states in the analysis by refining them or "de-emphasize" less important, similar states by combining them at any point during the assessment. These operations are especially useful when the network is large and local changes are desired which do not affect the rest of the network.

The refinment and coarsening operations have a dual relationship. In general, information is being removed in coarsening, and in refining information is being added to the network. Coarsening can "undo" refinement and if no information is lost in coarsening, a refinement can "undo" a coarsening. The refinement and coarsening operations developed in this paper can be thought of as "change of classification" operations in that they revise the classification (e.g., discretization) scheme for a given "axis" in a joint state space. A concrete example of this is the splitting of the state Vehicle into the two substates Tank and Truck as shown in Figure 4. This "change of classification" operation is only one type of state space modification. Another useful type of state space modification is the introduction or removal of classification axes. This is easily accomplished by adding or removing nodes from the network. For example, any new node that has no successors will not change any of the relationships between existing nodes.

The external and internal operations are closely related. For the external refinement operation one only needs to specify the splitting ratio between the new values. The conditional probabilities of the upper arcs and lower arcs are then generated automatically. The new arcs created by the external operation are always redundant and can be removed without any change to the joint distribution (see Figure 4). In the internal refinement operation one can specify more information than in the external operation since both the conditional probabilities of the upper arcs and lower arcs can be specified subject to certain constraints. Thus for refinement, the external operation is a special case of the internal operation.

However, this relationship is reversed for the coarsening operation—the internal operation is a special case of the external operation. The ability of the external operation to change the topology of the network allows any states to be coarsened whereas in the internal operation only if the constraints shown in equations (7) and (8) are satisfied can coarsening be performed. It is intuitive that only in such cases, the extra arcs create in the external operation are redundant. However since the main idea in coarsening is to reduce the state space, the introduction of new arcs, which is required in general, seems

**Conditional Prob for Node V**

Save  Normalize  Fill  Random

| M | A | U | N |
|---|---|---|---|
| A | 0.3 | 0.5 | 0.2 |
| B | 0.1 | 0.3 | 0.6 |

**Conditional Prob for Node F**

Save  Normalize  Fill  Random

| T | V | A | B | O |
|---|---|---|---|---|
| G | A | 0.4 | 0.5 | 0.1 |
| B | A | 0.1 | 0.5 | 0.4 |
| G | U | 0.6 | 0.3 | 0.1 |
| B | U | 0.4 | 0.2 | 0.4 |
| G | N | 0.1 | 0.1 | 0.8 |
| B | N | 0.2 | 0.2 | 0.6 |

**Conditional Prob for Node V**

Save  Normalize  Fill  Ra

| M | Y | N |
|---|---|---|
| A | 0.8 | 0.2 |
| B | 0.4 | 0.6 |

**Conditional Prob for Node F**

Save  Normalize  Fill  Random

| T | V | A | B | O |
|---|---|---|---|---|
| G | Y | 0.5375 | 0.3625 | 0.1 |
| B | Y | 0.30625 | 0.29375 | 0.4 |
| G | N | 0.1 | 0.1 | 0.8 |
| B | N | 0.2 | 0.2 | 0.6 |

**Posterior Table for Node M**

| A | B |
|---|---|
| 0.467 | 0.533 |

**Posterior Table for Node T**

| G | B |
|---|---|
| 0.702 | 0.298 |

**Posterior Table for Node V**

| Y | N |
|---|---|
| 0.732 | 0.268 |

**Posterior Table for Node F**

| A | B | O |
|---|---|---|
| 0.154 | 0.759 | 0.087 |

Figure 7: Approximate Coarsening with Internal Operation

**Conditional Prob for Node Vi**

Save  Normalize  Fill  Ra

| M | Y | N |
|---|---|---|
| A | 0.8 | 0.2 |
| B | 0.4 | 0.6 |

**Conditional Prob for Node F**

Save  Normalize  Fill  Random

| M | Vi | T | A | B | O |
|---|----|---|---|---|---|
| A | Y | G | 0.525 | 0.375 | 0.1 |
| B | Y | G | 0.55 | 0.35 | 0.1 |
| A | N | G | 0.1 | 0.1 | 0.8 |
| B | N | G | 0.1 | 0.1 | 0.8 |
| A | Y | B | 0.288 | 0.312 | 0.4 |
| B | Y | B | 0.325 | 0.275 | 0.4 |
| A | N | B | 0.2 | 0.2 | 0.6 |
| B | N | B | 0.2 | 0.2 | 0.6 |

**Posterior Table for Node M**

| A | B |
|---|---|
| 0.478 | 0.522 |

**Posterior Table for Node T**

| G | B |
|---|---|
| 0.701 | 0.299 |

**Posterior Table for Node Vi**

| Y | N |
|---|---|
| 0.733 | 0.267 |

**Posterior Table for Node F**

| A | B | O |
|---|---|---|
| 0.153 | 0.761 | 0.087 |

Figure 8: Coarsening with External Operation

counterproductive.

We feel an important application of this work is to the knowledge acquisition process. For Bayesian Networks, it is typical to first acquire, from an expert, the structure of a network. After the structure is determined, the state space of each node is acquired from the expert and lastly the probability distribution for each node is acquired. The structure is acquired first since this knowledge is the most robust cognitively. "Evidently, the notion of relevance and dependence are far more basic to human reasoning than the numerical values attached to probability judgements ... Once asserted, these dependency relationships should remain a part of the representation scheme, impervious to variations in numerical inputs."[1]. However this research shows there are definite constraints between structure, states, and probabilities.

Consider the example shown in Figures 7 and 8. Imagine that the structure in Figure 7 has been acquired and a decision is being made about the state space of node $V$. Consider the two possibilities: the state space of node $V$ is $Y$ and $N$ or the state space of node $V$ is $A$, $U$, and $N$. Imagine that we acquire the conditional probabilities for each possibility and assume the expert gives his "true" probabilities. Surprisingly, in general, the associated joint probability spaces for these two possibilities will be inconsistent. This leaves the issue of which state space possibility should be used. The intuitive answer is one should choose the state space which contains the "most information" but does not contain any "indistinguishable" states. In other words, a state space which is big enough but not too big! We call this the "maximumly distinguished" state space. The refining and coarsening operations introduced in the paper allow a formal definition of this term.

A "maximumly distinguished" state space is a state space which is both "irreducible" and "complete". An "irreducible" state space is one in which no coarsening operation can be performed without making the joint probability inconsistent (for the internal operation) or without introducing unremovable new arcs (for the external operation). Conceptually, a "complete" state space is one which contains enough distinctions to capture all the expert's knowledge. Stated in another way, a "complete" state space is a state space in which if any state is refined into substates, then the expert cannot distinguish between the substates. Formally then, a "complete" state space is one in which the expert probabilities on that state space can be reached by an internal coarsening operation on the expert probabilities of any more refined state.

This can be translated into broad guidelines for knowledge acquisition as related to network structure and state spaces. The knowledge engineer should first determine the structure of the network. Second, he should order the nodes such that the predecessors of a node are always before the node. He should then determine the state space and probability distribution of each node according to the order. This should be done by refining the state space of each node step-by-step, eliciting probabilities for each candidate state space. When the probabilities of a refined state space are "consistent" with a coarser state, prefer the coarser state. If a more refined state cannot be found after some search, then that acquisition of knowledge for that node can be considered complete.

## References

[1] R.A. Howard and J.E. Matheson. Influence diagrams. In R.A. Howard and J.E. Matheson, editors, *The Principles and Applications of Decision Analysis, vol. II*, Menlo Park: Strategic Decisions Group, 1981.

[2] J. Pearl. *Probabilistic Reasoning in Intelligent Systems: Networks of Plausible Inference*. Morgan Kaufmann Publishers, 1988.

[3] Ross D. Shachter. Intelligent probabilistic inference. In L.N. Kanal and J.F. Lemmer, editors, *Uncertainty in Artificial Intelligence*, Amsterdam: North-Holland, 1986.

[4] G. Shafer. *A Mathematical Theory of Evidence*. Princeton: Princeton University Press, 1976.

---

[1] [2], p. 79